\documentclass[]{spie}  %>>> use for US letter paper
\usepackage{amsmath,amsfonts,amssymb, bm}
\usepackage{graphicx}
\usepackage{booktabs}
\usepackage{enumitem}
\usepackage[colorlinks=true, allcolors=blue]{hyperref}
\usepackage{tabu}
\usepackage{wrapfig} % wrapping figures in text

\title{A Hypersensitive Breast Cancer Detector}

\author[1]{Stefano Pedemonte}
\author[1]{Brent Mombourquette}
\author[1]{Alexis Goh}
\author[1]{Trevor Tsue}
\author[1]{Aaron Long}
\author[1]{Sadanand Singh}
\author[1]{Thomas Paul Matthews}
\author[1]{Meet Shah}
\author[1]{Jason Su}

\affil[1]{Whiterabbit AI, Inc., Santa Clara, CA, USA}

\authorinfo{Further author information: (Send correspondence to whiterabbit.ai)\\whiterabbit.ai: E-mail: research@whiterabbit.ai}

\begin{document} 
\maketitle

%%%%%%%%%%%%%%%%%%%%%%%%%%%%%%%%%%%%%%%%%%%%%%%%%%%%%%%
%Prepare file as a PDF—
%For full consideration this file must include the:
%o Paper title
%o Authors
%o And the following supplemental information:
%▪ Description of purpose
%▪ Method(s)
%▪ Results
%▪ New or breakthrough work to be presented
%▪ Conclusions
%▪ Whether the work is being, or has been, submitted for publication or
%presentation elsewhere, and, if so, indicate how the submissions differ.
%▪ This file may contain supporting images/ tables /figures
%%%%%%%%%%%%%%%%%%%%%%%%%%%%%%%%%%%%%%%%%%%%%%%%%%%%%%%

\begin{abstract}
% REMOVED:
%Breast cancer is the second leading cause of cancer death for all women [CITE]. 
Early detection of breast cancer through screening mammography yields a 20-35\% increase in survival rate \cite{elmore2005screening}; 
%however, the Institute of Medicine \cite{national2005saving} and the American College of Radiology \cite{brogdon1990radiology} independently concluded that there are not enough radiologists to serve the growing population of women seeking screening mammography \cite{baxi2009breast}. 
however, there are not enough radiologists to serve the growing population of women seeking screening mammography \cite{baxi2009breast}.
Although commercial computer aided detection (CADe) software has been available to radiologists for decades, it has failed to improve the interpretation of full-field digital mammography (FFDM) images due to its low sensitivity over the spectrum of findings. In this work, we leverage a large set of FFDM images with loose bounding boxes of mammographically significant findings to train a deep learning detector with extreme sensitivity. Building upon work from the Hourglass architecture \cite{newell2016stacked}, we train a model that produces segmentation-like images with high spatial resolution, with the aim of producing 2D Gaussian blobs centered on ground-truth boxes. 
% MOST RECENT VERSION:
%Additionally, we replace the standard pixel-wise $L_2$ norm with a weakly-supervised loss designed to achieve high sensitivity that asymmetrically penalizes false positives and false negatives, ignoring the hardest false positives and utilizing a bank of reference blob shapes to make the loss invariant both to mis-alignments between the predicted and reference blobs and to the size of the predicted blobs. 
%Additionally, the standard pixel-wise $L_2$ norm loss is broken down into a background loss and a detection loss. By enforcing a restricted masking over images with no ground-truth annotations, the background loss downplays the effects of false positive findings which would otherwise lead to a conservative system with lower sensitivity. The detection loss softens the large impact of slight misalignments in pixel-wise operations by replacing the raw prediction with the nearest member of a bank of representative shapes. This latter loss emphasizes localization over exact segmentation. 
% Shorter version:
We replace the pixel-wise $L_2$ norm with a weak-supervision loss designed to achieve high sensitivity, asymmetrically penalizing false positives and false negatives while softening the noise of the loose bounding boxes by permitting a tolerance in misaligned predictions.
% REMOVED: Combined with an intensity peak-finding algorithm to compute coordinates, 
The resulting system achieves a sensitivity for malignant findings of 0.99 with only 4.8 false positive markers per image. When utilized in a CADe system, this model could enable a novel workflow where radiologists can focus their attention with trust on only the locations proposed by the model, expediting the interpretation process and bringing attention to potential findings that could otherwise have been missed. Due to its nearly perfect sensitivity, the proposed detector can also be used as a high-performance proposal generator in two-stage detection systems.
\end{abstract}
\begin{figure}
    \centering
    \includegraphics[width=1.0\textwidth]{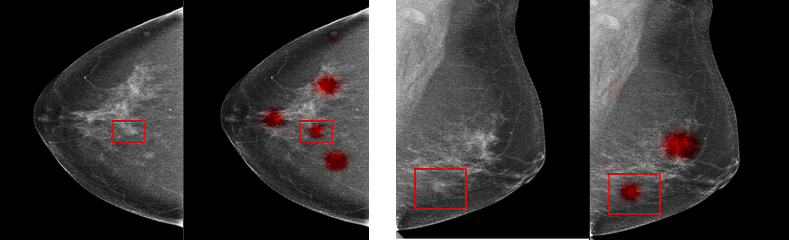}
    %\caption{Two examples of detections produced by the hypersensitive breast cancer detector. The red blobs represent the detector output. Malignant findings were marked with a bounding box by a radiologist. Left: example of caudal-cranial (CC) view with 3 false positives. Right: example of medio-lateral-oblique (MLO) view with one false positive.}
    \caption{Two examples of detections. Red blobs represent the detector output. Red boxes are Malignant findings marked by a radiologist. Left: 3 false positives. Right: One false positive.}
    \label{fig:examples_detections}
\end{figure}

\section{Purpose}
% Outline
%   breast cancer background
%   deep learning
%   our approach
%       highly sensitive detector

With nearly 269,000 new cases and 42,000 deaths each year, breast cancer is the most common and the second most deadly cancer for women \cite{cancerStatistics2019}. Breast cancer screening using full-field digital mammography (FFDM) has led to a reduction in deaths from this disease \cite{elmore2005screening}. Developed to aid radiologists in screening, traditional computer aided detection (CADe) highlighted suspicious regions in the image. However, it failed to achieve clinical utility due to its low sensitivity, high false positive rate, and reliance on hand-designed features \cite{lehman2015diagnostic,Baker2003}. 

Recently, convolutional neural networks (CNN) have achieved superhuman performance on many imaging tasks and have been adapted to FFDM to improve detection rates in cancer \cite{nyu:2019}. In particular, CNN detection models localize and classify different findings in images. However, most detection models are trained on well-labeled, balanced datasets with tight bounding boxes. These models can struggle to adapt to datasets without these qualities, such as mammography, where the cancer incidence rate is 0.51\% and most bounding boxes annotated in the routine clinical workflow only loosely encapsulate the finding \cite{Lehman2017}. Thus, we propose a new weakly-supervised loss function for a detector that marks malignant findings on mammograms that enables us to achieve extreme sensitivity and a false positive rate of less than a handful of marks per image.
% One approach for object detection involves localization through estimating an object's center coordinates \cite{newell2016stacked}. Such detectors have been developed mostly in the context of human pose estimation to localize body joints. A refreshingly simple and high performance solution was proposed by Newell et al. \cite{newell2016stacked}. They use a convolutional neural network trained to produced a Gaussian blob at the location of each joint. 
% NEW CUT: This Hourglass model is composed of a stack of U-Net networks \cite{ronneberger2015u} that progressively downsample and then upsample the features. 

\section{Methods}
\label{section:method}
\subsection{Hourglass Model}
Two principal approaches for object detection can be identified in the literature: 1 -- detectors that localize and predict the extent of objects by drawing bounding boxes around them \cite{He_2017_ICCV, liu2016ssd}, 2 -- detectors that localize objects by estimating their center coordinates \cite{newell2016stacked}. This second class of detectors has been developed mostly to localize body joints for human pose estimation. In this context, a refreshingly simple and high performance solution was proposed by Newell et al. \cite{newell2016stacked}. Their solution consists of a simple CNN trained to produced a Gaussian blob at the location of each joint. This CNN, named Hourglass, is composed of several U-Nets, which are stacks of residual modules that progressively downsample and then upsample the features \cite{ronneberger2015u}. Before each downsampling, skip connections are added across modules of identical resolution to facilitate gradient propagation. The key characteristic of the U-Net and Hourglass architectures is to enable the model to produce outputs with high spatial resolution while considering a wide receptive field for each output pixel. These end-to-end convolutional networks are not much different from VGG-style networks, but have the advantage of removing the trade-off between the size of the field-of-view and the resolution of the model output.

The authors of Hourglass achieve state-of-the-art performance on the body joint localization task by minimizing the $L_2$ norm between the network output and a stack of reference images, each representing a Gaussian blob located at a different joint location. In our approach for the localization of malignant lesions on mammograms, we maintain the Hourglass architecture but introduce a new loss that promotes high sensitivity.

\subsection{Blobs to coordinates}
\begin{figure}
    \centering
    \includegraphics[width=\textwidth]{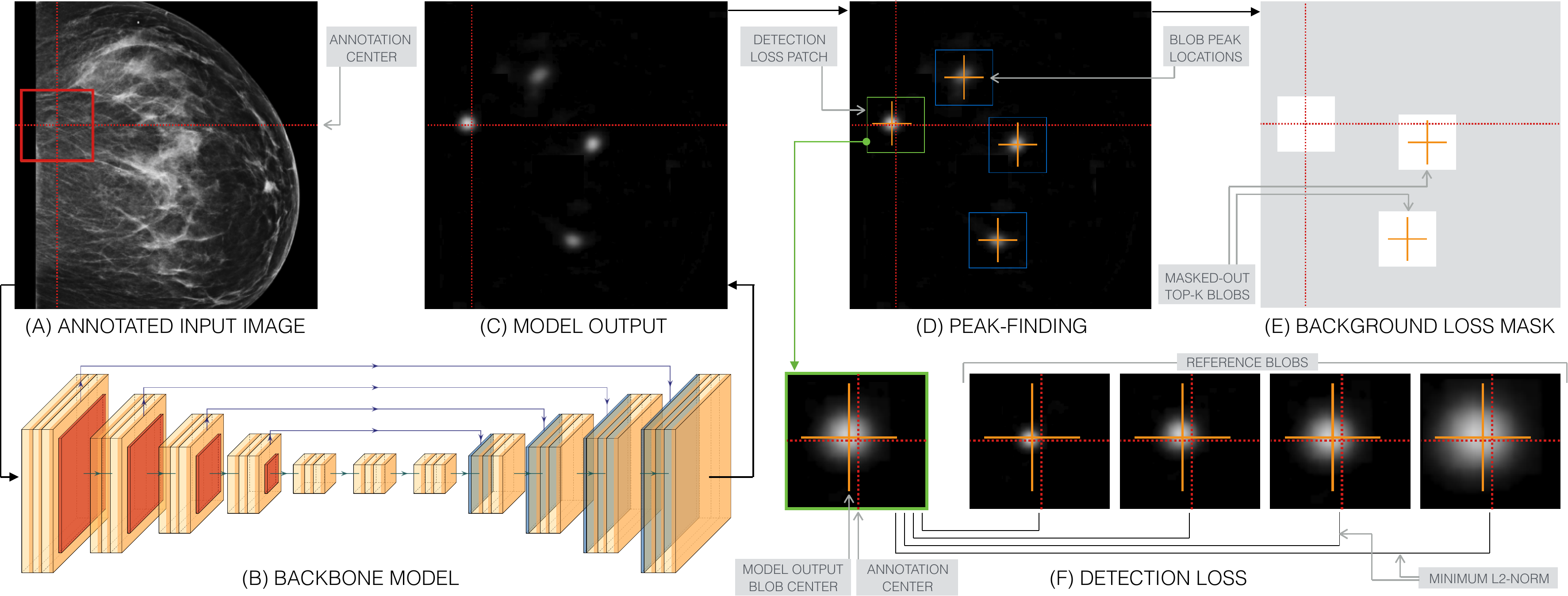}
    \caption{Hourglass network with loss that promotes high sensitivity}
    \label{fig:loss}
\end{figure}

In pose estimation, where there are a fixed set of points, the Hourglass is trained to produce a single blob output channel for each joint location in the form $\bm{O} \in \mathbb{R}^{N_\text{joints} \times N_x \times N_y}$. We modify the last layer of the network such that the output is $\bm{O} \in \mathbb{R}^{1 \times N_x \times N_y}$, capturing all potential findings in one channel. We then define a new loss that promotes high sensitivity and that allows for the prediction of any number of blobs at multiple locations. Conversion of $\bm{O}$ to an array of 2-dimensional coordinates is achieved by applying a simple peak-finding algorithm (See Figure \ref{fig:loss}-A-D). 
%In pose estimation, the Hourglass produces an output $\bm{O} \in \mathbb{R}^{N_\text{joints} \times N_x \times N_y}$ trained to generate a single blob for each feature map plane. We modify the network to produce an output $\bm{O} \in \mathbb{R}^{1 \times N_x \times N_y}$ feature map. We further define a loss that causes the network to learn to produce $k$ blobs at multiple locations. An additional model layer is then introduced, which converts the model output to an array of 2-dimensional coordinates by applying a simple peak-finding algorithm (See Figure \ref{fig:loss}-A-D).
% Removed for now
% This is implemented in GPU using the \texttt{maxpool} operator. The value at blobs centers is recorded as vector $I(n)$ and  utilized as a measure of confidence for each mark.

\subsection{An innovative loss that promotes sensitivity}
\label{section:loss}
In Hourglass, \cite{newell2016stacked} the loss is the $L_2$ norm of the difference between the predicted images and a stack of images containing 2-D Gaussian blobs centered on each joint annotation. We replace the $L_2$ norm of the  Hourglass model with one designed around four principles aimed at promoting sensitivity: 
\begin{enumerate}[label=(\alph*), topsep=1pt, itemsep=1pt,partopsep=1pt, parsep=1pt]
    \item Loss should remain unaffected by small mis-alignments between the predicted and ground truth locations.
    \item Loss should remain unaffected by size variations of the predicted blob.
    \item A small number of false positives per image is acceptable.
    \item False positive marks should be penalized less than false negative marks.
\end{enumerate}

\noindent The loss that we propose is composed of two terms: a detection loss $L_{\text{DET}}$ and a background loss $L_{\text{BG}}$. 
The detection loss operates on pixels surrounding the area of an annotation, measuring the similarity between the model output and a 2-D Gaussian blob. The background loss operates on pixels far away from annotations, measuring the similarity of the model output to zero. 

To calculate the detection loss, first a patch $\bm{O}_{\text{DET}}$ (marked in green in Figure \ref{fig:loss}-D,F) of a chosen fixed size centered on the ground truth annotation is extracted from the model output $\bm{O}$ (in a practical implementation, this operation happens in-place to enable propagation of the loss gradient). The detection loss is then calculated as the $L_2$-norm of the difference between $\bm{O}_{\text{DET}}$ and a reference patch $\bm{R}$. Following principles (a) and (b), we incorporate in the loss tolerances to errors in the location of annotations and to varying extent of the findings by constructing the reference patch $\bm{R}$ adaptively, as a function of the model output $\bm{O}_{\text{DET}}$. Invariance of the loss to small mis-alignments between the predicted and ground truth locations (a) is obtained by generating a reference 2-D Gaussian blob centered on the model's predicted blob (see Figure \ref{fig:loss}-F). Invariance of the loss to the predicted blob size (b) is obtained by comparing the model output $\bm{O}_{\text{DET}}$ with a bank of reference blobs $\bm{R}_i$ of different size $\bm{\sigma}_i$, with $i=1,\dots,N_{\text{RefBlobs}}$ (see Figure \ref{fig:loss}-F). Only the most similar blob (i.e. the blob that yields the smallest $L_2$-norm) contributes to the detection loss. Denoting with $\mathcal{N}$ a 2D Gaussian over patch coordinates $\bm{x}$:

\begin{equation}
\bm{x}_0 = \text{center\_of\_mass}(\bm{O}_{\text{DET}}),
\end{equation}
\begin{equation}
\bm{R}_i (\bm{x}) = \mathcal{N}\left(\bm{x}; \bm{x}_0, \bm{\sigma}_i\right),
\end{equation}
\begin{equation}
\bm{R} = \arg\min_i \lVert \bm{O}_{\text{DET}}-\bm{R}_i \rVert_2,
\end{equation}
\begin{equation}
L_{\text{DET}} = \lVert \bm{O}_{\text{DET}}-\bm{R}\rVert_2.
\end{equation}

\noindent The background loss enforces the model to produce values close to zero in areas far away from annotations, evaluating the $L_2$-norm of the model output in the region outside of the detector patch $\bm{O}_{\text{DET}}$ (see Figure \ref{fig:loss}-E). In order to promote sensitivity, we relax the effect of the background loss by allowing for a small number of false positives in each image (c). This is achieved by masking out patches of the model output centered on the top-$k$ highest confidence blobs generated by the model. Figure \ref{fig:loss}-E presents an example of background mask $\bm{M}(k)$ for $k=2$. 
\begin{equation}
L_{\text{BG}} = \lVert \bm{M}(k) \odot \bm{O} \rVert_2,
\end{equation}

\noindent where $\odot$ denotes element-wise multiplication. Finally, sensitivity is promoted by penalizing false positive marks less than false negative marks (d). This is achieved by down-weighting the  background loss by a factor $\omega \in [0,1]$, producing the final loss:
\begin{equation}
L = L_{\text{DET}} + \omega L_{\text{BG}}
\end{equation}

\subsection{Dataset}

\begin{wrapfigure}[15]{l}{.5\textwidth}
\centering
\vspace{-3mm}
\begin{tabular}{llll}
    %\vspace*{1mm}\\
    \toprule
     %\cmidrule(l){2-4} \cmidrule(l){5-6} \cmidrule(l){7-8}
     ~ & Train & Val & Test  \\
     \midrule
     Patients & 49748 & 6214 & 6194\\
     Exams & 158159 & 19872 & 19556\\
     Images & 665805 & 84135 & 82450 \\
     \midrule
     Annotations & & & \\
     \ \ \ \ Benign & 11930 & 1556 & 1380 \\
     \ \ \ \ High Risk & 891 & 124 & 76 \\
     \ \ \ \ Malignant & 2699 & 331 & 293 \\
    %  \ \ \ \ Other & 12821 & 1680 & 1456 \\
     \midrule
     Normal Images & 653232 & 82520 & 81009 \\
     % & & & \\
     \bottomrule
     \hspace*{1mm}\\
\end{tabular}
\caption{Image-level statistics for training (Train), validation (Val), and testing (Test) datasets and annotation counts.}
\label{tab:data}
\end{wrapfigure}

A total of 197,587 screening mammography exams (832,390 FFDM images) spanning 62,156 patients were collected from an academic medical center in the United States. The exams were interpreted by one of 11 radiologists with breast imaging experience ranging from 2 to 30 years. Annotations were collected as part of the routine clinical work flow. The intended clinical use case of these annotations did not require precise segmentations nor rigorous definitions of the physical extent of a finding. Therefore, the tightness of an annotation to the finding's boundary varies from case to case. The data encompasses all types of mammographically significant findings that could be encountered in a screening setting except for breast implants which were excluded from both training and evaluation. Screening exams were associated with subsequent biopsy events by clinical staff for regulatory compliance purposes through dedicated mammography reporting software (Magview 7.1, Burtonsville, Maryland). This provided a structured way to directly link annotations on screening exams to pathology results from biopsies. Pathology cell type information was mapped to the labels of benign, high risk, or malignant by a fellowship trained breast imaging radiologist.

The images were classified into four classes: (1) normal, no suspicious tissue was found by a radiologist, (2) benign, benign tissue was found during screening or biopsy, (3) high risk, tissue likely to develop into cancer was found during biopsy, (4) malignant, malignant tissue was found during biopsy. During training, the model was trained to identify high risk (3) and malignant (4) as the positive cases with normal (1) and benign (2) as the negative cases.  Our analysis focused on the detection of biopsy proven malignant findings. Therefore, though we considered high risk findings as positive examples in training to promote sensitivity, during evaluation sensitivity was evaluated with malignant (4) as the positive class and all others as the negative class. Patients were randomly selected for model training, validation, or testing according to a 80:10:10 split. These splits were on the patient level, so there are no overlapping images, exams, or patients in the different datasets. Training, hyperparameter tuning, and model selection were completed using only the training and validation sets. The final performance was evaluated once on the test set after all models had been frozen. Statistics of the three data sets are reported in Table \ref{tab:data}.

%\begin{wraptable}{r}{0.55\textwidth}%{9cm}
%    \centering
%    \begin{tabular}{llll}
%        %\vspace*{1mm}\\
%        \toprule
%         %\cmidrule(l){2-4} \cmidrule(l){5-6} \cmidrule(l){7-8}
%         ~ & Train & Val & Test  \\
%         \midrule
%         Patients & 49748 & 6214 & 6194\\
%         Exams & 158159 & 19872 & 19556\\
%         Images & 665805 & 84135 & 82450 \\
%         \midrule
%         Benign annotations & 11930 & 1556 & 1380  \\
%         High Risk annotations & 891 & 124 & 76 \\
%         Malignant annotations & 2699 & 331 & 293 \\
%         No finding images & 653232 & 82520 & 81009 \\
%         % & & & \\
%         \bottomrule
%         \hspace*{1mm}\\
%    \end{tabular}
%    \caption{Image-level statistics and annotation counts of datasets for training (Train), validation (Val), and testing (Test).}
%    \label{tab:data}
%\end{wraptable}

\subsection{Training}

The model was trained on FFDM images resized to $3840 \times 3840$ then downsampled by a factor of 2.5 in each dimension. Training occurred for 80 epochs using the Adam optimizer with a learning rate of 1e-4 and background weight $\omega=0.01$. One epoch consisted of 8,000 images evenly sampled among the three annotation types and the set of images without findings. Owing to the fully-convolutional architecture, the size of the input image can be different for each batch. Thus, in order to speed-up experimentation, we trained the model in two phases. First, we pre-trained the model on patches of size $1280 \times 1280$ centered on image annotations, downsampled by 2.5, and then fine-tuned it on the whole-image input with size $3840 \times 3840$, downsampled by 2.5. 

\section{Results}

\begin{figure}
    \centering
    \begin{tabular}{lcclcl}
        \toprule
      \multicolumn{1}{c}{Method}    &   Sensitivity & FPI & Data Type & Malignant/Total Cases & Dataset\\
        \midrule
        Malich \textit{et al.} (2001) \cite{Malich2001}     &   0.900       &   1.3     &   M, C        & 150/150 & Private\\
        Petrick \textit{et al.} (2002) \cite{Petrick2002}   &   0.870       &   1.5     &   M           & 156/156 & Private\\
        Baker \textit{et al.} (2003) \cite{Baker2003}       &   0.380       &   0.7     &   D           & 45/45 & Private\\
        % Hu \textit{et al.} (2011) \cite{Hu2011}             &   0.913       &   0.7     &   M, D, A     & & miniMIAS\\
        Dhungel \textit{et al.} (2015) \cite{Dhungel2015}   &   0.75        &   4.8     &   M           & 40/40 & DDSM-BCRP\\
        Morra \textit{et al.} (2015) \cite{Morra2015}       &   0.890       &   2.7     &   M, C        & 123/175 & Private\\
        Ribli \textit{et al.} (2018) \cite{Ribli2018}       &   0.9         &   0.3  &   M, C, D, A     & 115/115 & INbreast\\
        % Teuwen \textit{et al.}  (2018) \cite{Teuwen2018}    &   0.969       &   3.6     &   M, D, A     & 2883/7196 & Private (40\% is test)\\
        % Moor \textit{et al.} (2018) \cite{Moor2018}         &   0.94        &   7.9     &   M, D, A     & 2883/7196 & Private (40\% is test)\\
        Teuwen \textit{et al.}  (2018) \cite{Teuwen2018}    &   0.969       &   3.6     &   M, D, A     & 1153/2878* & Private\\
        Moor \textit{et al.} (2018) \cite{Moor2018}         &   0.94        &   7.9     &   M, D, A     & 1153/2878* & Private\\
        Agarwal \textit{et al.} (2019) \cite{Agarwal2019}   &   0.980       &   1.7     &   M           & 211/223 & DDSM + INbreast\\
        Ours                                                &   0.990       &   4.8     &   M, C, D, A  & 168/19556 & Private\\
        \bottomrule
        \hspace*{1mm}\\
    \end{tabular}
    \caption{Sensitivity and false positives per image (FPI) for different CADe systems. M = Masses, C = Microcalcifications, D = Architectural Distortions, A = Asymmetries. * Approximation (exact numbers not given)}
    \label{tab:sensitivity}
\end{figure}

\begin{figure}
    \vspace{-8mm}
    \centering
    \includegraphics[width=0.65\textwidth]{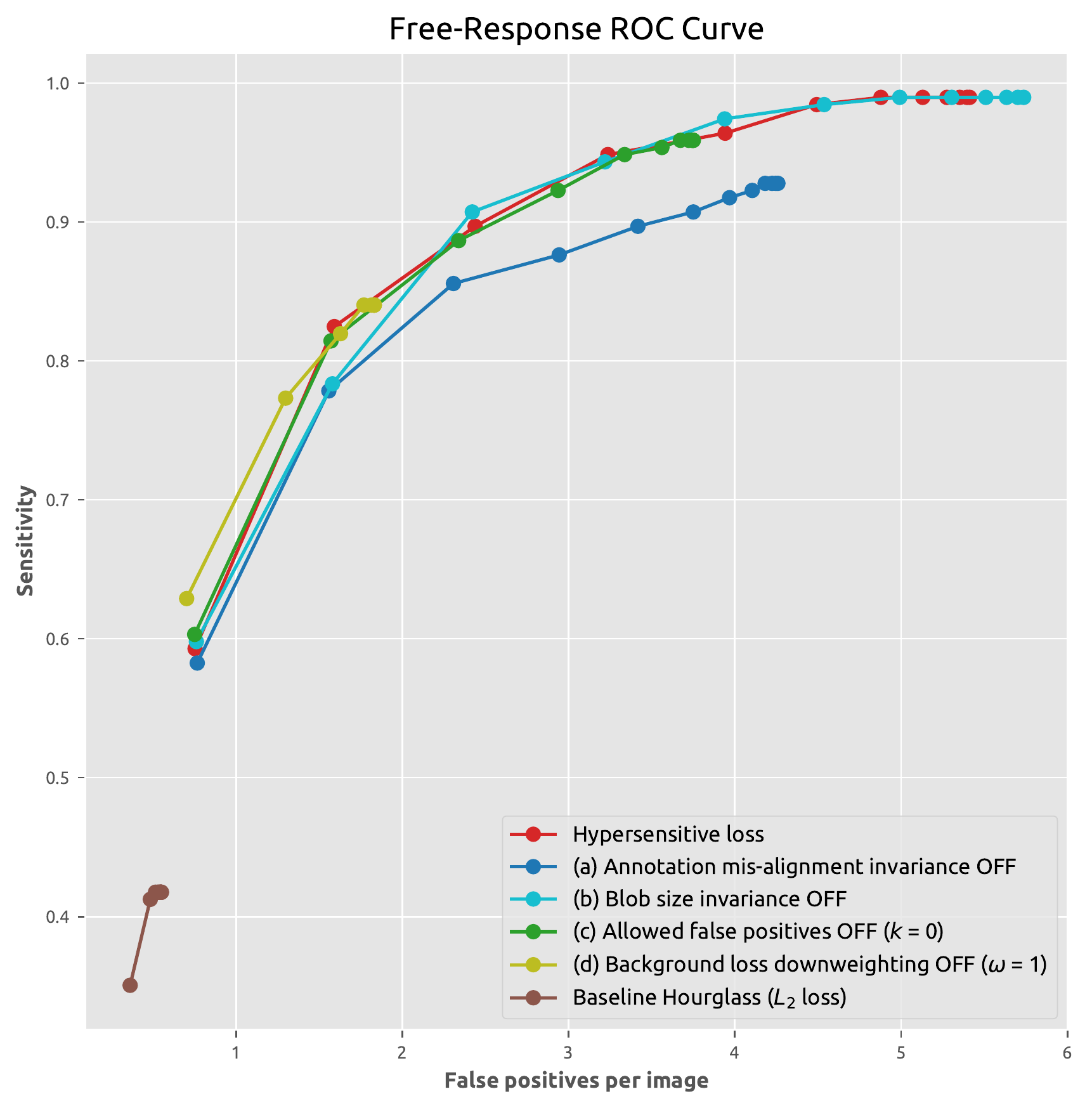}
    \caption{Effect of removing the four aspects of the loss on sensitivity and false positives per image (FPI). The plots represent sensitivity as a function of the average number of false positive marks per image for increasing values of a threshold applied to the predicted Gaussian blob intensity. All aspects increase the sensitivity: (a) invariance to blob alignment, (b) invariance to blob size, (c) accepting top-$k$ false positives per image ($k=3$), (d) down-weighting the background loss ($\omega=0.01$). For several of the curves (baseline, (a), (c), (d)), the sensitivity does not reach a value greater than 0.96 even when all blobs are accepted, confirming the importance of each component of the loss. Blob size invariance (b) has the effect of reducing the false positive rate while maintaining maximum sensitivity.}
    \label{fig:sensitivity}
\end{figure}

The hypersensitive loss detector achieves a sensitivity for malignancies of 0.99, generating only 4.8 false negative marks per image. The Hourglass model with the original $L_2$ loss only achieves 0.42 sensitivity, so standard detection loss functions struggle on this challenging detection problem and highly imbalanced dataset.  Additionally, we compare our approach to other CADe software in Table \ref{tab:sensitivity}. Our approach is the only method that both evaluates on a highly imbalanced dataset that reflects the natural distribution (0.86\% malignancy case occurrence) and detects masses, microcalcifications, architectural distortions, and asymmetries. These two features are crucial for a clinically-functional model, as they reflect what a model would encounter in an actual screening population, where only 0.51\% of exams have cancer and exams can have any type of lesion \cite{Lehman2017}. Additionally, we experiment in Figure \ref{fig:sensitivity} to demonstrate the positive effect on sensitivity of each of the four innovative components of the loss described in Section \ref{section:loss}.

\section{Conclusions}
%The new loss function that we devised to train hypersensitive detectors can be used as a drop-in replacement in the Hourglass and U-Net and in newer architectures such as HR-Net \cite{sun2019deep}.

% ORIGINAL ABSTRACT VERSION:
%The nearly perfect sensitivity achieved by the proposed model could allow radiologists to safely ignore image regions not highlighted by the algorithm. We believe that this is an important step forward to improve, simplify, and substantially expedite radiologists' interpretations of mammograms. 
%This new loss function can be used to train a hypersensitive detector with any segmentation-based network architecture.

This work presents a novel loss function for training mammography CADe models. By considering the specific properties of both the problem domain and the data, design and optimization of this loss produce a hypersensitive detector with an acceptable false positive rate. Moreover, this loss function, though inspired by mammography CADe, has components directly applicable to many image-based detection tasks. Data with large variances in both segmentation quality and physical extent of findings to be detected are exceedingly common especially in the medical imaging domain. The proposed loss function makes no assumptions about the underlying model and can be used with any segmentation-based network architecture.

The low sensitivity of existing CADe systems limits their applicability in the radiologists' interpretation workflow. Since malignant findings can be missed by a low-sensitivity CADe system, the human reader cannot rely on software generated annotations. Therefore existing CADe systems are utilized as second readers to bring back the radiologist's attention to image areas that the algorithm considers suspicious. However, unlike other CADe systems, our model performs with a nearly perfect sensitivity and an acceptable rate of false positives on a dataset that reflects the natural distribution, with a 0.86\% cancer occurrence rate of all the different types of lesions: masses, microcalcifications, architectural distortions, and asymmetries. Thus, this nearly perfect sensitivity to malignant findings enabled by the new loss function described in this work could potentially allow radiologists to safely ignore regions of the images not highlighted by the algorithm. We believe that these are important steps forward for improving, simplifying, and substantially expediting radiologists' interpretations of mammograms. 
\section{Future Work}
The high sensitivity of this detection model allows the detection of nearly all malignant findings in all images, making it an ideal proposal-generator for a two-stage detection system \cite{NIPS2015_5638}. In the first stage, this detection model identifies the suspicious regions of each full image, ideally capturing all the malignant lesions with high sensitivity. In the second stage, we can train a classifier on image patches from the proposals generated by the first stage. This second model solely inputs smaller patches and is trained specifically on these proposals, which are the most suspicious and challenging image areas to classify. Such a two stage model could increase the specificity while maintaining the high level of sensitivity.
\section{Disclosure}
This work has not been submitted to any journal or conference for publication or presentation consideration.

\bibliography{report} % bibliography data in report.bib
\bibliographystyle{spiebib} % makes bibtex use spiebib.bst
\end{document}